\documentclass[letterpaper]{article} 
\usepackage{aaai2026}  
\usepackage{times}  
\usepackage{helvet}  
\usepackage{courier}  
\usepackage[hyphens]{url}  
\usepackage{graphicx} 
\usepackage{natbib}  
\usepackage{caption} 
\title{Unsupervised Feature Selection Through Group Discovery}
\author {
    Shira Lifshitz\textsuperscript{\rm 1},
    Ofir Lindenbaum\textsuperscript{\rm 2},
    Gal Mishne\textsuperscript{\rm 3},
    Ron Meir\textsuperscript{\rm 1},
    Hadas Benisty\textsuperscript{\rm 1}
}
\affiliations {
    \textsuperscript{\rm 1}Technion -- Israel Institute of Technology\\
    \textsuperscript{\rm 2}Bar Ilan University\\
    \textsuperscript{\rm 3}University of California San Diego\\
    shiralif@campus.technion.ac.il, ofirlin@gmail.com, gmishne@ucsd.edu, rmeir@ee.technion.ac.il, hadasbe@technion.ac.il
}

\usepackage{booktabs}   
\usepackage{amsmath} 
\DeclareMathOperator*{\argmin}{arg\,min}
\usepackage{amssymb,amsfonts}
\usepackage{pifont}   
\newcommand{\tick}{\ding{51}}         
\newcommand{\selg}[1]{\tick\,\textbf{#1}}
\usepackage{multirow}
\usepackage{pdfpages}

\begin{document}

\maketitle

\begin{abstract}
Unsupervised feature selection (FS) is essential for high-dimensional learning tasks where labels are not available. It helps reduce noise, improve generalization, and enhance interpretability. However, most existing unsupervised FS methods evaluate features in isolation, even though informative signals often emerge from groups of related features. For example, adjacent pixels, functionally connected brain regions, or correlated financial indicators tend to act together, making independent evaluation suboptimal. Although some methods attempt to capture group structure, they typically rely on predefined partitions or label supervision, limiting their applicability. We propose GroupFS, an end-to-end, fully differentiable framework that jointly discovers latent feature groups and selects the most informative groups among them, without relying on fixed a priori groups or label supervision. GroupFS enforces Laplacian smoothness on both feature and sample graphs and applies a group sparsity regularizer to learn a compact, structured representation. Across nine benchmarks spanning images, tabular data, and biological datasets, GroupFS consistently outperforms state-of-the-art unsupervised FS in clustering and selects groups of features that align with meaningful patterns.  
\end{abstract}


\section{Introduction}
Modern machine-learning systems routinely handle datasets with thousands to millions of features. Such high-dimensional data arise in neuroscience, finance, and computer vision~\cite{fan2010selective, donoho2000high}. However, many of the observed features are nuisance, i.e., uninformative or noisy, and they obscure latent structure, inflate computational cost, and degrade generalization.  
Feature selection (FS) tackles this problem by retaining only the most relevant features, thereby discarding nuisance dimensions, reducing computational cost, and boosting downstream performance, e.g., clustering accuracy~\cite{guyon2003introduction}.  
Because FS preserves the original measurements, the results remain interpretable and can enable domain-specific insights.  
In real-world applications such as neuroimaging, FS can lower acquisition costs by focusing on task-relevant regions, thus saving time or enabling higher resolution. Similarly, in domains like behavioral research, feature acquisition (e.g., questionnaires) can be expensive or burdensome, making efficient feature selection especially valuable.

While many FS methods are supervised, even without labels, a well-chosen subset of features can uncover latent structure~\cite{li2017feature}. Yet selecting that subset is a complex combinatorial problem, and the challenge is amplified in the unsupervised setting where there are no labels to guide the selection process. Since obtaining annotations often requires costly expert effort, robust unsupervised FS is both challenging and essential~\cite{solorio2020review, li2024exploring}.

Classical FS methods can be categorized into three families.  
\emph{Filter} methods assign scores to features using model‑agnostic criteria such as mutual information or graph smoothness~\cite{he2005laplacian, battiti1994using}.  
\emph{Wrapper} methods search over subsets by repeatedly training a model, incurring high computational cost~\cite{kohavi1997wrappers}.  
\emph{Embedded} methods impose sparsity while training the model itself, e.g.\ LASSO~\cite{tibshirani1996regression} or stochastic‑gating networks~\cite{yamada2020feature, sristi2023contextual}.  
Most of these approaches, however, score features independently and ignore relationships among them.

Real-world features often ``act together": spatially adjacent pixels, temporally co-varying sensors, or functionally coupled genes. Such relationships suggest that grouping features into meaningful subsets and selecting at the group level, rather than individually, can boost performance and provide clearer scientific insights. Existing group-aware FS methods either assume groups are known a priori or rely on supervision to form them~\cite{you2023mgagr, imrie2022composite}. However, in many applications, the group structure is unknown, and fixing groups in advance can bias selection. Jointly discovering groups, selecting which ones are informative and rejecting the rest, without labels, remains an open problem, referred to as unsupervised group feature selection.

In this paper, we address this gap by introducing \textbf{GroupFS}. It is a fully differentiable, end-to-end framework that simultaneously learns feature groups and selects the informative ones in a purely unsupervised manner. Our approach constructs two graphs: one over the sample space and another over the feature space, enforcing Laplacian smoothness on both. A feature-grouping and gating mechanism, guided by sparse regularization, dynamically discovers relevant feature groups.

Our main contributions are as follows:
\begin{itemize}
    \item We introduce GroupFS, the first end-to-end, fully differentiable framework for unsupervised feature selection that jointly discovers latent feature groups and selects informative groups from them.
    \item GroupFS automatically learns latent feature groups without relying on predefined partitions or supervision, thereby broadening its applicability to unlabeled, real-world data.
    \item Extensive experiments on diverse synthetic and real-world datasets demonstrate that GroupFS consistently outperforms state-of-the-art unsupervised FS baselines in clustering accuracy and identifies meaningful feature groups.
\end{itemize}

\section{Related Work}

\noindent \textbf{Unsupervised FS.} One line of research addresses the unsupervised FS problem by constructing a sample graph and selecting features that vary smoothly over the data manifold~\cite{he2005laplacian, cai2010unsupervised, lindenbaum2021differentiable, miao2022graph,luo2024adaptive}. Autoencoder-based methods offer an alternative, ranking features by their contribution to reconstruction loss~\cite{abid2019concrete,svirsky2024interpretable}. 
However, reconstruction objectives do not necessarily promote features that capture the relationships among samples, which are essential for downstream tasks such as clustering.

\noindent \textbf{Group FS.} Other work seeks to exploit feature groups, but most approaches assume the groups are fixed a priori, using heuristics or domain knowledge to define them~\cite{you2023mgagr, zaharieva2017unsupervised, perera2020group, wang2017exploiting, park2024feature}. Although effective in special cases, predefined groups limit adaptability and can introduce bias. A more flexible strategy is to learn groups during training. \citet{imrie2022composite} take a step in this direction by jointly inferring group structure and training a classifier, but they rely on label supervision.~\citet{sristi2022disc} proposes a spectral approach to select groups of features; however, they assume a setting of differentiating between two or more given datasets.

In contrast to these, we jointly discover feature groups and select the informative ones without any supervision, allowing structure to emerge directly from the data.

\section{Preliminaries}
\subsection{Graphs and Spectral Analysis}
\label{sec:graphs}
Let $X=[\mathbf{x}_1,\ldots,\mathbf{x}_N]^{\top}\in\mathbb{R}^{N\times d}$ be a data matrix, where row $i$ (sample $i$) is $\mathbf{x}_i = X_{i:}\in\mathbb{R}^d$, and column $k$ (feature $k$) is $\mathbf{x}^{(k)} = X_{:k}\in\mathbb{R}^N$.
We assume the data lies on a low-dimensional manifold and capture its local geometry using an undirected, weighted graph \(G = (V, E, W)\)~\cite{von2007tutorial, ng2001spectral}.  
Pairwise affinities are defined using the self-tuning kernel~\cite{zelnik2004self} as
\begin{equation}
\label{eq:w_self}
W_{ij} = \exp\!\left(-\frac{\|\mathbf{x}_i - \mathbf{x}_j\|_2^2}{\gamma_i \gamma_j}\right),
\end{equation}
where \(\gamma_i\) is the distance from \(\mathbf{x}_i\) to its \(K\)-th nearest neighbor.  
This sample-dependent scaling adapts to local density, improving robustness in heterogeneous data.  
The degree matrix is defined as \(D = \operatorname{diag}(d_1, \dots, d_N)\), where \(d_i = \sum_j W_{ij}\).  
Two standard graph operators are the normalized graph Laplacian \(L_{\text{sym}} = I - D^{-1/2} W D^{-1/2}\) and the random walk matrix \(P = D^{-1} W\). Though both are used in spectral analysis, they differ in interpretation: eigenvectors corresponding to low eigenvalues of \(L_{\text{sym}}\) capture smooth, low-frequency variations on the graph, while those associated with high eigenvalues of \(P\) capture similar directions. The matrix power \(P^t\) represents transition probabilities after \(t\) steps of the random walk~\cite{spielman2025sagt}.

\paragraph{Laplacian Score (LS).} 
\citet{he2005laplacian} leveraged a sample graph structure for feature selection by ranking individual features based on their alignment with the graph's smoothest modes. Features whose values vary minimally across strongly connected samples (i.e., along high‑weight edges) are preferred.
For a feature vector \(\mathbf{x}^{(k)} \in \mathbb{R}^{N}\), the Laplacian Score is
\begin{equation*}
\mathrm{LS}(\mathbf{x}^{(k)})
   \;=\;{(\mathbf{x}^{(k)})}^{\top}L_{\mathrm{sym}}\mathbf{x}^{(k)}
   \;=\;\sum_{i=1}^{N}\lambda_{i}\,\langle\mathbf{v}_{i},\mathbf{x}^{(k)}\rangle^{2},
\end{equation*}
where \(\{(\lambda_{i}, \mathbf{v}_{i})\}_{i=1}^{N}\) are the eigenpairs of \(L_{\mathrm{sym}}\).  
A smaller score indicates that the feature varies smoothly over the sample manifold and is therefore considered more informative. Using the matrix trace operator $\operatorname{tr}$, the total Laplacian score becomes
\begin{equation*}
\sum_{k=1}^{d} (\mathbf{x}^{(k)})^\top L_{\mathrm{sym}} \mathbf{x}^{(k)}
    = \operatorname{tr}\!\left(X^{\top} L_{\mathrm{sym}} X\right).
\end{equation*}

\begin{figure*}[t]
  \centering
  \includegraphics[width=0.8\linewidth]{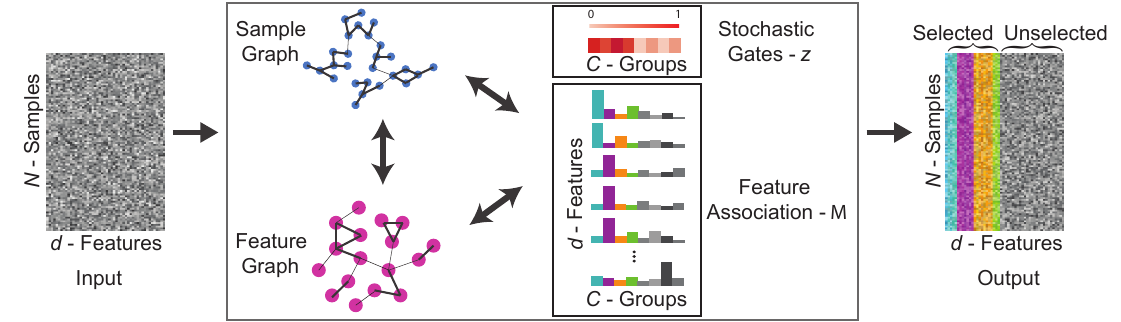}
\caption{Illustration: GroupFS learns feature-to-group associations, enforces smoothness on the feature graph, infers the importance of each group, and reconstructs a smoother sample-similarity graph.}
\label{fig:illustration}
\end{figure*}

\subsection{Gumbel-Softmax}
\label{sec:gumbel}
The Gumbel-Softmax~\cite{jang2016categorical}, also known as the Concrete distribution~\cite{maddison2016concrete}, provides a differentiable approximation to categorical sampling. It relaxes the discrete one-hot vector into a continuous distribution over \( C \) classes. Given class probabilities \(\boldsymbol{\pi} = [\pi_1, \pi_2, \dots, \pi_C]\) and a temperature $T > 0$, we draw i.i.d.\ Gumbel noise variables $g_c \sim \mathrm{Gumbel}(0,1)$ and compute
\begin{equation*}
m_c = \frac{\exp\!\bigl((\log \pi_c + g_c)/T\bigr)}
            {\sum_{h=1}^{C} \exp\!\bigl((\log \pi_h + g_h)/T\bigr)}.
\label{eq:gumbel_softmax}
\end{equation*}
As $T \to 0$, the distribution becomes increasingly peaked, and $\mathbf{m}\in \mathbb{R}^C$ approaches a one-hot sample.
The reparameterization trick~\cite{Kingma2013Auto} enables gradients to propagate through the sampling process, allowing Gumbel-Softmax to be trained using standard gradient-based optimizers \cite{battash2024revisiting}.

\subsection{Stochastic Gates}
\label{sec:stg}
The stochastic gates method~\cite{yamada2020feature,jana2023support} provides a differentiable mechanism for feature selection by learning relaxed Bernoulli gates for the features. Each input feature \(\mathbf{x}^{(k)}\), for \(k \in \{1, \dots, d\}\), is multiplied by a stochastic gate (STG):
$z_k = \max(0, \min(1, \mu_k + \varepsilon_k))$
where \(\varepsilon_k \sim \mathcal{N}(0, \sigma^2)\) and \(\mu_k\) are learnable parameters. This clipped Gaussian variable produces continuous approximations of binary gates. The expected number of selected features is
\begin{equation*}
\mathbb{E}[\|\mathbf{z}\|_0] = \sum_{k=1}^{d} \mathbb{P}(z_k > 0) = \sum_{k=1}^{d} \Phi\left(\frac{\mu_k}{\sigma}\right),
\end{equation*}
where \(\Phi(\cdot)\) denotes the standard Gaussian CDF and \(\sigma\) is the fixed gate noise. This relaxation enables training with standard gradient-based optimizers while implicitly encouraging sparsity through \(\ell_0\)-style regularization.

\section{GroupFS}

We tackle unsupervised group FS without assuming prior knowledge about the groups. Instead, we simultaneously learn feature groups and select the most relevant ones, yielding a compact and interpretable model (see Figure~\ref{fig:illustration}).

\paragraph{Problem setup.}
Let \(X \in \mathbb{R}^{N \times d}\) be a data matrix with \(N\) samples and \(d\) raw features.  
We assume the features can be partitioned into \(C\) latent groups \(\{\mathcal{G}_1, \dots, \mathcal{G}_C\}\), which are unknown a priori.  
Our method uncovers latent groups of related features, selects specific groups needed to preserve the data’s intrinsic geometry, and learns a low-dimensional embedding that faithfully reflects that geometry. Our model is guided by a composite loss function consisting of three components:
\begin{itemize}
    \item \textbf{Sample-wise smoothness (\(\mathcal{L}_s\))}: Encourages feature values to vary smoothly across the sample manifold, promoting gradual transitions between nearby data points.
    \item \textbf{Feature-wise smoothness ($\mathcal{L}_f$)}: Encourages consistent group assignments among high‑affinity neighbors on the feature graph.
    \item \textbf{Group sparsity (\(\mathcal{L}_{\text{reg}}\))}: Promotes the selection of a small number of informative feature groups, resulting in a compact and interpretable model.
\end{itemize}

\paragraph{Overall Loss.} Our objective combines the three components:
\begin{equation*}
\mathcal{L} = \mathcal{L}_s + \lambda_1 \cdot \mathcal{L}_f + \lambda_2 \cdot \mathcal{L}_{\text{reg}},
\label{eq:total_loss}
\end{equation*}
where \(\lambda_1\) and \(\lambda_2\) weigh the relative importance of each term. This unified, end-to-end framework integrates differentiable grouping, stochastic gating, and Laplacian smoothness to discover informative feature groups while filtering out irrelevant or noisy features in an unsupervised setting. We describe each component below.

\subsection{Sample-wise Smoothness Loss \(\mathcal{L}_{s}\)}
\label{sec:sample-wise}

\paragraph{Feature Association.}
Given a batch \(X_B \in \mathbb{R}^{B \times d}\), we learn a feature-to-group assignment matrix \(M \in \mathbb{R}^{d \times C}\) using the Gumbel-Softmax trick (see Sec.~\ref{sec:gumbel}).  
Each \emph{row} \(M_{i,:}\) encodes the soft membership of feature \(i\) across the \(C\) latent groups,
\begin{equation*}
M_{ij}
  = \frac{\exp\!\bigl((\log \pi_{ij} + g_{ij}) / T\bigr)}
         {\sum_{k=1}^{C} \exp\!\bigl((\log \pi_{ik} + g_{ik}) / T\bigr)},
\end{equation*}
where \(\pi_{ij}\) are learnable logits, \(g_{ij} \sim \mathrm{Gumbel}(0, 1)\) is i.i.d.\ noise, and \(T\) is a temperature parameter annealed during training.  
As \(T \to 0\), each row approaches a one-hot vector, effectively assigning feature \(i\) to a single group.  
Since \(M\) is learned jointly with the group-importance gates (see following sub-section), the model can discover meaningful groupings directly from the data.

\paragraph{Group Importance.}
Following the STG formulation (see Sec.~\ref{sec:stg}), we attach a stochastic gate $z_j$ to each feature group, reducing the number of learnable gating parameters from \(d\) (features) to \(C\) (groups). We select features by sorting the groups according to their gate means and retaining those from the top-ranked groups. To compute feature-level weights, we aggregate the gated group assignments:
\begin{equation*}
  \hat z_i = \sum_{j=1}^{C} M_{ij} \cdot z_j,\quad
  i \in \{1, \dots, d\}, \quad j\in \{1, \dots, C\}.
\end{equation*}
Intuitively, \(\hat z_i\) measures the importance of feature \(i\) by averaging over its soft group memberships weighted by each group’s gate.
Broadcasting \(\hat{\mathbf{z}}\) across the batch yields \(\hat{Z} \in \mathbb{R}^{B \times d}\) with identical rows. We apply these gates to the input using element-wise multiplication: 
\(\widetilde{X} = X_B \odot \hat{Z}\), effectively masking out less important features.

\paragraph{Smoothness Objective.}
We construct a dense random walk matrix \(P_{\widetilde{X}}\) (Sec.~\ref{sec:graphs}) over the batch-masked input \(\widetilde{X}\) at each iteration, using the affinity matrix defined in eq.~\eqref{eq:w_self}.  
To promote smooth variation over the gated sample manifold, we use
\begin{equation*}
\mathcal{L}_s
  = -\frac{1}{B\,d}\,
    \operatorname{tr}\left(\widetilde X^{\top} P_{\widetilde X}^{t} \widetilde X\right).
\end{equation*}

Here, \(P_{\widetilde X}^{t}\) denotes the \(t\)-step diffusion operator, i.e., the \(t\)-th power of the random-walk matrix. Maximizing the trace aligns the retained features with the graph’s low-frequency directions. This encourages the model to assign higher importance to feature groups whose values vary smoothly across the sample manifold.

$\mathcal{L}_s$ combines the group assignments $M$ with their gate importances $\textbf{z}$ to select a subset of groups, and then rebuilds the sample graph using only these features. On this newly constructed graph, the retained features vary smoothly, turning the raw data into a cleaner and more informative representation.

\subsection{Feature-wise Smoothness Loss \(\mathcal{L}_{f}\)}

This term encourages the learned cluster assignments to vary smoothly across features while remaining mutually distinct.  
We first embed each feature into a $C$-dimensional space:
$F = M\,Q \in \mathbb{R}^{d \times C}$
where \(M \in \mathbb{R}^{d \times C}\) is the soft assignment matrix from Section~\ref{sec:sample-wise}, and \(Q \in \mathbb{R}^{C \times C}\) is a trainable linear projection that allows interactions between clusters.

\paragraph{Smoothness on the feature graph.}
Analogous to the sample graph in Section~\ref{sec:graphs}, we construct a feature similarity graph, where nodes represent features, and compute its normalized Laplacian \(L_{\mathrm{feat}} \in \mathbb{R}^{d \times d}\).  
The term
\(
\operatorname{tr}\left(F^{\top} L_{\mathrm{feat}} F\right)
\)
penalizes rapid changes of \(F\) across similar features, promoting alignment with the graph's low-frequency directions.

\begin{figure}[t]
    \centering
    \includegraphics[width=0.8\linewidth]{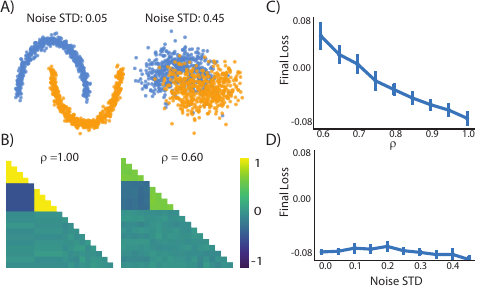} 
\caption{\textbf{Two-moons synthetic data.}  
(A) 2D visualization of the dataset under low and high Gaussian noise levels (\(\text{STD} = 0.05\) and \(\text{STD} = 0.45\)).  
(B) Feature correlation matrices (\(20 \times 20\), lower triangle) with two levels of correlation strength (\(\rho = 1.00\) and \(\rho = 0.60\)).  
(C) Final training loss as a function of correlation strength \(\rho\), showing lower loss for stronger correlations.  
(D) Final training loss as a function of noise standard deviation, showing robustness to moderate sample-level noise.  
Results in (C,D) are averaged over 10 random seeds; error bars denote standard error.}
    \label{fig:synthetic}
\end{figure}

\paragraph{Orthogonality regularization.}
To ensure diverse, non-redundant cluster embeddings, analogous to orthogonal eigenvectors in spectral clustering, we add an orthogonality penalty using the Frobenius norm:
\(\lVert F^{\top} F - I \rVert_F^2\). Combining both objectives yields:
\begin{equation*}
\mathcal{L}_{f} =
\frac{1}{d\,C} \left[
      \operatorname{tr}\left(F^{\top} L_{\mathrm{feat}} F\right)
    + \beta\cdot\lVert F^{\top} F - I \rVert_F^2
\right],
\end{equation*}
where $\beta$ is a hyper-parameter that weights the orthogonality penalty.
Following each update step, we center the columns of \(F\) to have zero mean and renormalize them to unit \(\ell_2\)-norm.  
This avoids convergence to trivial solutions such as constant vectors or the zero vector. 

Intuitively, $\mathcal{L}_f$ term encourages similar features (or highly connected nodes in the feature graph) to have similar group assignments, ensuring that the learned groups respect the underlying structure of the feature space.

\subsection{Group Sparsity Loss \(\mathcal{L}_{\mathrm{reg}}\)}

To encourage selection of the most informative groups, we penalize the expected number of active gates, weighted by each group’s relative size.  Using the activation probability for STG gates (Sec.~\ref{sec:stg}), we define the regularization term
\begin{equation*}
\mathcal{L}_{\mathrm{reg}} = \frac{1}{C}\sum_{j=1}^{C} \mathbb{P}(z_j > 0)\cdot \frac{1}{d}\sum_{i=1}^d M_{ij},
\end{equation*}
which increases with both the likelihood that group \(j\) is active and the proportion of features assigned to it.  
Minimizing \(\mathcal{L}_{\mathrm{reg}}\) therefore encourages compactness and sparsity by keeping fewer and smaller groups active.

\section{Experiments}

We evaluate \textbf{GroupFS} across three complementary settings:

\begin{enumerate}
    \item \textbf{Synthetic data.} 
    We construct a synthetic dataset with features partitioned into known groups.  
    This setup allows us to (i) verify whether GroupFS correctly recovers and selects the true groups, and (ii) study the effects of hyperparameters and intrinsic data properties.

    \item \textbf{Real-world data.} 
    To assess whether explicit feature grouping enhances or hinders FS, we compare GroupFS to state-of-the-art baselines across nine widely used datasets from image and biological domains. 

    \item \textbf{Interpretability.}  
    We demonstrate that GroupFS discovers meaningful feature groups that align with domain knowledge.
\end{enumerate}

\subsection{Implementation Details}
\label{sec:implement_details}
The model’s learnable parameters are:
(i) $d \times C$ logits of the Gumbel-Softmax assignment matrix $M$;
(ii) $C$ gate means $\{\mu_j\}$ for the STG-based group importances; and
(iii) $C \times C$ transformation matrix $Q$. A heuristic for selecting the number of groups $C$ 
is described in Appendix~\ref{app:choose_C_appendix}.

\paragraph{Initialization.}
Gates are initialized to $\mu_j = 0.5$ (an unbiased prior) following
\cite{yamada2020feature}.
We warm-start the logits using spectral clustering assignments based on
$L_{\mathrm{sym}}$~\cite{von2007tutorial}:
for a feature $i$ assigned to cluster $j^\star$, we set
\[
\log(\pi_{ij}) =
\begin{cases}
\Delta & \text{if } j = j^\star \\[2pt]
0      & \text{otherwise}
\end{cases} ,\;
\Delta = \log\!\left(\tfrac{p_{\text{main}}}{p_{\text{rest}}}\right),
\]
with $p_{\text{rest}} = \tfrac{1 - p_{\text{main}}}{C - 1}$ and $p_{\text{main}} = 0.7$ in all experiments. We initialize $Q$ as a random orthonormal matrix and scale each row inversely to the feature-cluster sizes estimated from the spectral-clustering logits initialization, ensuring balanced influence on $F$. Note that these initial group assignments are not fixed; they are gradually overwritten during training as the model adapts the gates and groupings to minimize the total loss.
Further information regarding hyperparameters is detailed in App.~\ref{app:hparam_groupFS}.

\begin{figure*}[t]
  \centering
  \includegraphics[width=0.80\textwidth]{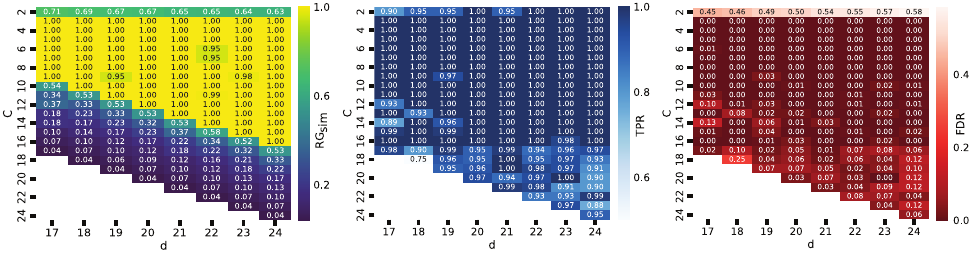}
    
    \caption{\textbf{Two-moons: Effect of feature dimension $d$ and group count $C$.}
    Mean $RG_{\text{sim}}$ TPR and FDR of the best-loss model over 10 random seeds. Complementary std results are in App.~\ref{app:exp_synthetic}.}

  \label{fig:grid_2moons}
\end{figure*}

\subsection{Synthetic Data}
We construct a 20-dimensional synthetic dataset by extending the classic \emph{two-moons} dataset (Fig.~\ref{fig:synthetic}A). Features 1-5 are noisy linear transformations of the moons' first coordinate, and features 6-10 of the second, each generated as $x' = \sqrt{\rho}\, x + \sqrt{1 - \rho}\, \epsilon$, where $x$ is an original coordinate and $\epsilon \sim \mathcal{N}(0,1)$. The remaining features (11-20) are i.i.d.\ Gaussian noise with zero mean and unit variance. The correlation strength $\rho \in [0.6, 1]$ controls how tightly each group follows its base coordinate. Figure~\ref{fig:synthetic}B shows two examples for the sample correlation matrices (lower triangle) across features for $\rho{=}1.0$ and $\rho{=}0.6$ (a detailed description regarding the construction of the synthetic dataset is in App.~\ref{app:data_details_synthetic}). The goal is to identify $G_1{=}\{1{:}5\}$ and $G_2{=}\{6{:}10\}$ as two separated groups, while the rest (11-20) are assigned to other groups, activate the gates attached to $G_1$ and $G_2$, and deactivate the other gates attached to the rest. Unless noted otherwise, we use $\rho{=}0.95$, additive Gaussian noise with std.\ $0.05$, 500 training epochs, batch size 100, and determine group count via the heuristic described in App.~\ref{app:choose_C_appendix}.

\paragraph{Effect of correlation strength.}  
Fig.~\ref{fig:synthetic}C shows how the final loss varies with correlation strength $\rho$. For each $\rho$, we select the best model from a grid search over $\lambda_1$ and $\lambda_2$, choosing the combination that yields the lowest final loss averaged over 10 runs with different seeds (details in App.~\ref{app:hparam_groupFS}). As $\rho$ increases, the loss decreases, reaching a minimum at $\rho{=}1$.  
This trend aligns with the more transparent block structure in the correlation matrices (Fig.~\ref{fig:synthetic}B), confirming that GroupFS favors stronger intra-group coherence.

\paragraph{Effect of additive noise.}
In Fig.~\ref{fig:synthetic}D, we plot the final loss as a function of the noise std. This curve remains essentially flat, indicating that increasing the standard deviation of the additive Gaussian noise up to ~0.45 has little to no effect on the final loss.  
This suggests that GroupFS is robust to moderate sample-level noise, a desirable property in real-world applications where such noise is common.

Across all tested noise levels and $\rho$, the model isolates the informative groups $\{1{:}5\}$ and $\{6{:}10\}$ while the ten noisy dimensions are placed in unselected clusters.

\begin{table*}[t]
\centering
\setlength{\tabcolsep}{1mm} 
\begin{small}               
\begin{tabular}{lccccccccc c} 
\toprule
Dataset & ALL & LS & MCFS & CAE & DUFS & MGAGR & CompFS & GroupFS & \#Feat & Dim/Samp/Class \\
\midrule
ALLAML            & $65.1\!\pm\!8.4$ & $\mathbf{70.6\!\pm\!1.4}$ & $69.7\!\pm\!4.4$ & $67.4\!\pm\!3.2$ & $66.1\!\pm\!5.2$ & $66.4\!\pm\!4.8$ & $57.2\!\pm\!6.3$ & $\mathbf{70.6\!\pm\!1.4}$ & 274 & 7129 / 72 / 2 \\
Lung500   & $86.1\!\pm\!10.9$ &  $83.4\!\pm\!4.3$ &$84.8\!\pm\!7.1$ & $91.3\!\pm\!6.7$ & $88.6\!\pm\!7.3$  &  $82.0\!\pm\!8.9$ & $81.3\!\pm\!8.8$ & $\mathbf{93.0\!\pm\!6.8}$ & 234 & 500 / 56 / 4 \\

METABRIC          & $65.7\!\pm\!6.4$ & $64.2\!\pm\!5.0$ & $65.3\!\pm\!5.7$ & $70.5\!\pm\!8.2$ & $60.4\!\pm\!8.1$ & $\mathbf{70.6\!\pm\!6.0}$ & $63.8\!\pm\!4.0$ & $68.0\!\pm\!3.2$ & 226 & 489 / 1904 / 2 \\
HeartDisease     & $82.5\!\pm\!0.7$ & $81.9\!\pm\!1.3$ & $82.6\!\pm\!0.4$ & $78.5\!\pm\!1.2$ &  $77.1\!\pm\!8.1$& $75.1\!\pm\!7.4$ & $82.0\!\pm\!0.4$ & $\mathbf{83.1\!\pm\!0.5}$ & 10   & 13 / 297 / 2 \\
Yale              & $\mathbf{46.6\!\pm\!3.5}$ & $41.5\!\pm\!2.2$ & $37.8\!\pm\!3.5$ & $46.0\!\pm\!3.9$ & $43.3\!\pm\!4.9$ & $37.6\!\pm\!5.0$ & $44.1\!\pm\!6.1$ & $42.1\!\pm\!1.4$ & 341 & 1024 / 165 / 15 \\
AR10P             & $24.7\!\pm\!4.5$ & $22.5\!\pm\!1.1$ & $22.7\!\pm\!2.8$ & $19.6\!\pm\!2.1$ & $23.7\!\pm\!1.9$ & $24.9\!\pm\!2.9$ & $24.7\!\pm\!3.2$ & $\mathbf{32.5\!\pm\!4.1}$ & 362 & 2400 / 130 / 10 \\
PIE10P            & $29.0\!\pm\!2.7$ & $22.8\!\pm\!1.0$ & $34.0\!\pm\!2.0$ & $24.4\!\pm\!1.5$ & $35.0\!\pm\!3.6$ & $34.4\!\pm\!2.8$ & $29.0\!\pm\!1.8$ & $\mathbf{38.4\!\pm\!2.5}$ & 49  & 2420 / 210 / 10 \\
NMNIST 3-8    & $72.6\!\pm\!7.6$ & $77.3\!\pm\!1.0$ & $67.2\!\pm\!0.2$ & $57.1\!\pm\!0.3$ & $51.1\!\pm\!0.7$ & $52.3\!\pm\!10.7$ & $64.6\!\pm\!0.9$ & $\mathbf{83.3\!\pm\!0.1}$ & 51  & 784 / 1000 / 2 \\
NMNIST        & $\mathbf{49.5\!\pm\!2.7}$ & $46.1\!\pm\!2.4$ & $48.9\!\pm\!6.9$ & $44.1\!\pm\!3.0$ & $20.1\!\pm\!5.9$ & -- & $45.9\!\pm\!1.4$ & $48.9\!\pm\!2.7$ & 184 & 784 / 12000 / 10 \\
\bottomrule
\end{tabular}
\end{small}

\caption{\textbf{Scenario 1 - Fixed budget, unsupervised setting.} $k$-means accuracy (mean $\pm$ std over 10 runs). All methods use the same feature budget (\#Feat). The last column shows the original feature dimension, sample count, and number of classes. Bold marks the best score per dataset.}
\label{tab:fixed_budget}
\end{table*}

\paragraph{Effect of feature and group numbers.}
We assess performance along two axes: (i)~\emph{feature grouping} and (ii)~\emph{feature selection}.  
In this controlled setting, the true group structure is known ($G_1{=}\{1{:}5\}$, $G_2{=}\{6{:}10\}$), enabling quantitative evaluation. We adapt the Group Similarity metric~\cite{imrie2022composite} to compute \emph{Relevant-Group Similarity} (\textbf{RG\textsubscript{sim}}).  
Let $\mathcal{G} = \{G_1, G_2\}$ denote the ground-truth groups and $\widehat{\mathcal{G}} = \{\hat G_1,\dots,\hat G_C\}$ the predicted groups. We retain only predicted groups that overlap with at least one informative group: $
\widetilde{\mathcal{G}}
= \bigl\{\hat G_j \mid \hat G_j\cap G_1\neq\varnothing
                          \;\text{or}\;
                          \hat G_j\cap G_2\neq\varnothing\bigr\}$.
Then we define
\begin{equation*}
RG_{\text{sim}}
  = \frac{1}{\max\bigl(|\mathcal{G}|, |\widetilde{\mathcal{G}}|\bigr)}
    \sum_{i=1}^{2}
      \max_{\hat G_j\in\widetilde{\mathcal{G}}}
      \mathcal{J}\bigl(G_i,\hat G_j\bigr),
\label{eq:rg_sim}
\end{equation*}
where
\(
\mathcal{J}(A,B)=\lvert A\cap B\rvert/\lvert A\cup B\rvert
\)
is the Jaccard index.  
This score lies in $[0,1]$ and achieves a value of 1 only when both informative groups are perfectly recovered.

For feature selection, we report two common metrics.  
\emph{True Positive Rate} (\textbf{TPR}) is the fraction of informative features $\{1{:}10\}$ that are selected (preferred: TPR${=}1$).  
The \emph{False Discovery Rate} (\textbf{FDR}) is the fraction of selected features from the noise set $\{11{:}20\}$ (preferred: FDR${=}0$).

We vary the total number of features $d$ (the last $d{-}10$ are nuisance) and the number of groups $C$, then evaluate the best-loss model over 10 runs with the three metrics (hyperparameters in App.~\ref{app:hparam_groupFS}).  Importantly, ground‑truth groups are used only to compute evaluation metrics; the model is trained without access to this information. We retain the top-ranked groups by gate mean until at least 10 features are covered. The overall results are summarized in Fig.~\ref{fig:grid_2moons}.
We note that the effective number of groups that is required to fully separate signal and noise is: $C{=}2{+}(d{-}10)$, one group per informative feature cluster plus one per nuisance feature. Indeed, we observe that for $C \leq 2{+}(d{-}10)$, the model nearly always achieves $RG_{\text{sim}}{=}1$, TPR${=}1$ and FDR${=}0$, indicating it cleanly recovers both informative groups and assigns noise to separate groups.  
An exception is $C{=}2$: with too few groups, noise features are merged with informative ones, lowering $RG_{\text{sim}}$. TPR stays high, but FDR rises due to unwanted noise selection.  
When $C > 2{+}(d{-}10)$, performance remains strong in TPR and FDR, but $RG_{\text{sim}}$ gradually declines. In this case, the model breaks informative clusters into smaller groups, a reasonable outcome given the surplus of available groups.

Overall, GroupFS performs well across a range of \(C\) values, as long as \(C\) is neither too small to separate informative from noisy features, nor too large to over-fragment the groups. In practice, setting \(C\) slightly above the expected number of informative groups is effective, especially for high-dimensional data.

\begin{table*}[t]
\centering
\setlength{\tabcolsep}{1mm}      
\begin{small}                    
\begin{tabular}{lccccccc}
\toprule
Dataset & LS & MCFS & CAE & DUFS & MGAGR & CompFS & GroupFS \\
\midrule
ALLAML        & 72.2\,(200) & 71.8\,(200) & 70.4\,(100) & 71.5\,(400) & 66.3\,(200) & 67.4\,(100) & \textbf{72.8\,(302)} \\
Lung500       & 82.2\,(200) & 87.5\,(200) & 92.3\,(400) & 95.0\,(50)  & 93.0\,(100) & 94.5\,(100) & \textbf{96.1\,(361)} \\
METABRIC      & 68.0\,(200) & 69.3\,(100) & 72.8\,(50)  & 72.4\,(100) & 71.6\,(200) & 73.2\,(50)  & \textbf{73.4\,(159)} \\
HeartDisease  & 81.9\,(10)  & 82.7\,(8)   & 82.8\,(8)   & 83.5\,(8)   & \textbf{84.3\,(10)} & \textbf{84.3\,(8)} & 83.1\,(10) \\
Yale          & 43.0\,(400) & 45.0\,(100) & 45.6\,(400) & 42.6\,(200) & 40.7\,(100) & 46.7\,(50)  & \textbf{48.3\,(398)} \\
AR10P         & 32.9\,(50)  & 29.5\,(100) & 30.1\,(100) & 34.2\,(200) & 32.6\,(50)  & 29.8\,(50)  & \textbf{34.7\,(363)} \\
PIE10P        & 26.3\,(400) & 34.1\,(100) & 26.6\,(400) & \textbf{42.1\,(50)} & 36.0\,(50) & 29.1\,(50) & 40.8\,(370) \\
NMNIST 3–8    & 76.6\,(200) & 68.1\,(400) & 76.0\,(400) & 80.2\,(200) & 77.4\,(400) & 80.4\,(100) & \textbf{84.1\,(288)} \\
NMNIST        & 49.4\,(400) & 50.4\,(200) & 50.5\,(400) & 42.0\,(400) & -- & \textbf{55.0\,(200)} & 48.9\,(184) \\
\bottomrule
\end{tabular}
\end{small}
\caption{\textbf{Scenario 2 - Adaptive budget, accuracy‑guided setting.} $k$‑means accuracy (mean over 10 runs). Each method selects its own feature budget (numbers in parentheses). Bold marks the best score per dataset. See App.~\ref{app:exp_real_data} for mean$\pm$std results.}

\label{tab:changed_budget_main}
\end{table*}

\subsection{Real-world Data}
\label{sec:benchmarks}

Our evaluation spans nine widely used datasets drawn from two domains (see Table \ref{tab:fixed_budget} for sizes).
The biomedical set: ALLAML~\cite{golub1999molecular}, Lung500~\cite{lee2010biclustering}, METABRIC~\cite{pereira2016somatic, curtis2012genomic}, and HeartDisease~\cite{heart_disease_45} contains gene-expression or clinical profiles.
The vision set includes Yale~\cite{cai2007learning}, AR10P~\cite{martinez1998ar}, PIE10P~\cite{sim2002cmu}, and two noisy MNIST variants~\cite{larochelle2007empirical, lecun2002gradient}, all of which utilize image data. We adopt a random-background version of noisy MNIST (NMNIST), and a “3-8” subset comprising 500 images of digit 3 and 500 images of digit 8 (NMNIST 3-8).
All datasets are $z$-scored feature-wise.
Full download links and details appear in App.~\ref{app:data_details_real}. We compare GroupFS against a diverse set of FS methods:
\begin{itemize}
    \item \textbf{LS} and \textbf{MCFS}: Classical graph-based selectors~\cite{he2005laplacian, cai2010unsupervised}.
    \item \textbf{CAE}: Concrete Autoencoder based on the Gumbel-Softmax relaxation~\cite{abid2019concrete}.
    \item \textbf{DUFS}: A stochastic gating approach that learns a sample graph during training~\cite{lindenbaum2021differentiable}.
    \item \textbf{MGAGR}: A recent unsupervised method that leverages pre-defined feature groups~\cite{you2023mgagr}.
    \item \textbf{CompFS}: A supervised baseline that jointly learns feature groups and a classifier~\cite{imrie2022composite}.
    \item \textbf{ALL}: A trivial baseline using all features.
\end{itemize}

We assess FS quality via $k$-means clustering, with $k$ set to the number of ground-truth classes.  
To reduce sensitivity to initialization, we run $k$-means ten times with different seeds and report mean clustering accuracy \(\pm\) standard deviation. To ensure a fair comparison, we adapt each baseline as needed:  
Graph-based methods (LS, MCFS) use the self-tuning kernel from eq.~\eqref{eq:w_self} to avoid manual bandwidth tuning.  
CAE and CompFS are trained with a 90/10 train-validation split, and the best model is chosen based on the lowest reconstruction loss (CAE) or the highest accuracy (CompFS) on the validation set.  
Because CompFS lacks a global feature budget, we aggregate all learned scores and retain the top-ranked features.
MGAGR follows the authors’ recommended grouping but is skipped on NMNIST due to impractical runtime. For full hyperparameter settings and implementation details, see App. \ref{app:hparam_baselines}.

\noindent \textbf{Scenario 1: Fixed budget, unsupervised model choice (Table~\ref{tab:fixed_budget}).}  
All methods use the same feature budget as GroupFS (i.e., number of selected features). We determine this budget by gradually adding feature groups, ordered by the mean of their gates, and choosing the number of groups (and corresponding number of features) that yields a local maximum in model accuracy. For all models, hyperparameters are set based on the lowest loss, without supervision (i.e., without using labels), except for CompFS, which uses validation accuracy due to its supervised nature.  
Our GroupFS ranks first or tied for first on 6 out of 9 datasets, outperforming the next-best method by an average of +3.84\%.  
On two of the remaining three datasets, GroupFS still ranks in the top three.
Notably, Yale and NMNIST appear especially challenging for feature selection, as using all features yields the best results on both.

\noindent \textbf{Scenario 2: Adaptive budget, accuracy-guided model choice (Table~\ref{tab:changed_budget_main}).}
We ran each baseline with multiple feature budgets: \(\{50, 100, 200, 400\}\) (or \(\{2, 4, 8, 10\}\) for the Heart Disease dataset) and reported the result with the highest \(k\)-means accuracy.
Since GroupFS naturally outputs variable-sized groups, we
retain the parameter choice that achieves the best accuracy while using no more than 400 features (or 10 for the Heart Disease dataset). In this setting, GroupFS achieves the highest accuracy on 6 out of 9 datasets, demonstrating strong performance even when methods are allowed to adapt their feature count.

\begin{figure}[t]
  \centering
  \includegraphics[width=0.75\linewidth]{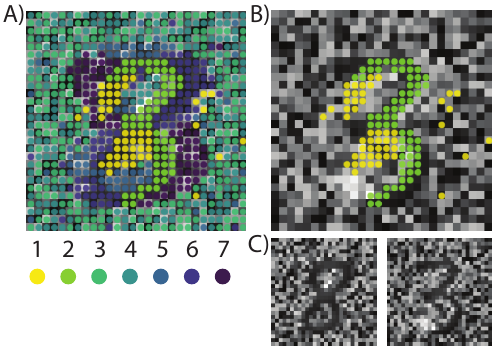}
\caption{\textbf{GroupFS on NMNIST (3 vs.\ 8).}  
(A) Pixel groups discovered by GroupFS, colored by group ID and ranked by importance (1 = highest, 7 = lowest).  
(B) The top two groups align with class-relevant regions.  
(C) Noisy image examples of digits `8' and `3'.}\label{fig:noisy_mnist_groups}
\end{figure}

\subsection{Interpretability Experiment}
\label{sec:qualitative}
We evaluate the interpretability benefits of GroupFS on two datasets from distinct domains: vision and education. In both cases, our goal is to highlight how unsupervised group discovery reveals meaningful feature groups aligned with domain knowledge.
The first is the NMNIST 3-8 subset, and the second is the UCI Student Performance dataset~\cite{student_performance_320}, with 395 samples and 30 features covering academic, demographic, and behavioral variables. For the UCI dataset, we focus on predicting math exam pass/fail outcomes (a detailed description of this dataset is App. \ref{app:data_details_real}).

\noindent \textbf{NMNIST 3-8.}  
GroupFS discovers seven spatially coherent pixel groups, visualized in Figure~\ref{fig:noisy_mnist_groups}.  
Pixels are colored by group, with the legend ranking importance from 1 (most important) to 7 (least).  
Although learned with no supervision, group 1 (yellow) highlights regions that differentiate 3s from 8s, such as the upper-left loop of 8s, which is typically absent in 3s.  
Lower-ranked groups correspond mainly to background pixels with little discriminative value.  
This shows that GroupFS segments the image into functionally meaningful regions that are spatially localized and consistent across the dataset.

\noindent \textbf{Student Performance.} For this tabular dataset, GroupFS discovers seven interpretable feature groups. Taking the top three groups leads to clustering accuracy of $61.3\pm2.6\%$. The highest-ranked group includes features related to alcohol consumption (daily and weekly). The second group includes features associated with motivation, such as the number of school absences, past academic failures, romantic relationships, and intention to pursue higher education. The third group relates to parents, including their education and the mother’s job. Each group displays strong semantic coherence, reinforcing the idea that GroupFS can uncover meaningful structure even in unlabeled tabular data. A full breakdown of feature rankings for this and other baselines appears in App.~\ref{app:exp_studentperf}.

\section{Conclusion}
We address unsupervised group feature selection, which involves the discovery of informative groups of features in an entirely data-driven manner, without relying on labels or prior knowledge. We introduce \textbf{GroupFS}, an end-to-end differentiable framework that simultaneously learns group assignments and sparsely selects them. Across a variety of image, tabular, and biomedical datasets, GroupFS matches or surpasses state-of-the-art unsupervised feature-selection methods on downstream clustering tasks under both fixed and variable feature budgets.  Qualitative inspection further shows that the discovered groups align with meaningful domain structure.
GroupFS has limitations: the sample and feature graphs rely on Euclidean distances, which can misrepresent data on curved, non-Euclidean manifolds. It also learns a single global notion of group importance, overlooking condition, or time-dependent relevance, where both groups and their importance may evolve. Future work will include smooth, differentiable manifold-aware distances and a dynamic formulation with condition-adaptive grouping and ranking.
Overall, GroupFS provides a practical step toward combining feature‑structure discovery with sparse selection in a purely unsupervised setting, and it can be used modularly as a building block for downstream learning tasks.

\section*{Acknowledgements}
This research was partially supported by the NSF (CCF-2403452), by the ISF (2418/24) and (1693/22), and by the Skillman chair (RM). OL was supported by the MOST grant No. 0007341.

\bibliography{aaai2026}
\clearpage 
\appendix

\section{Data Details}

\subsection{Synthetic Data.}
\label{app:data_details_synthetic}

We evaluate GroupFS on a synthetic benchmark derived from the classic two moons dataset, where each of the $N$ samples lies in $\mathbb{R}^2$ and is labeled as one of two classes. Let $x_n = (x_n^{(1)}, x_n^{(2)}) \in \mathbb{R}^2$ denote the 2D coordinates of sample $n \in \{1, \dots, N\}$. To simulate a higher-dimensional setting with latent group structure, we extend each sample to $d$ features as follows:
\begin{itemize}
    \item Features 1-5 are noisy, linearly correlated versions of the first coordinate $x_n^{(1)}$.
    \item Features 6-10 are similarly derived from the second coordinate $x_n^{(2)}$.
    \item Features 11–$d$ are i.i.d.\ Gaussian noise.
\end{itemize}

Let $x_n^{(*)}$ denote either coordinate $x_n^{(1)}$ or $x_n^{(2)}$. Each of the first 10 features is generated as
\[
    y_n^{(i)} = \sqrt{\rho} \cdot x_n^{(*)} + \sqrt{1 - \rho} \cdot \epsilon_n, \quad i = 1, \ldots, 5,
\]
where:
\begin{itemize}
    \item $\rho \in [0.6, 1]$ controls the correlation strength,
    \item $\epsilon_n \sim \mathcal{N}(0, 1)$ are i.i.d.\ Gaussian noise variables,
    \item $x_n^{(*)}$ is standardized to have zero mean and unit variance.
\end{itemize}
This design yields two distinct correlation regimes.
\begin{itemize}
    \item Within-base-coordinate features:
          For any pair derived from the same coordinate,
          \[
              \operatorname{Corr}\bigl(y_n^{(i)},y_n^{(j)}\bigr)=\rho.
          \]
    \item Cross-base-coordinate features:
          For features derived from different coordinates,
          \[
              \operatorname{Corr}\bigl(y_n^{(i)},y_n^{(j)}\bigr)
              =\rho\,\operatorname{Corr}\!\bigl(x^{(1)},x^{(2)}\bigr).
          \]
          In the standardized two-moons data we measure
          $\operatorname{Corr}(x^{(1)},x^{(2)})\approx -0.45$.
        \item Features that involve any purely noisy coordinate are essentially uncorrelated, i.e.\ $\operatorname{Corr}\approx 0$. 
\end{itemize}

This setup provides ground-truth group structure, enabling a controlled evaluation of both grouping and selection performance.

\subsection{Real-World Data.}
\label{app:data_details_real}

The datasets ALLAML~\cite{golub1999molecular}, Yale~\cite{cai2007learning}, AR10P~\cite{martinez1998ar} and PIE10P~\cite{sim2002cmu} are publicly available through the scikit-feature collection\footnote{\url{https://jundongl.github.io/scikit-feature/datasets.html}}. Six datasets are not included in that collection:

\begin{itemize}
    \item \textbf{Heart Disease (UCI)}\footnote{\url{https://archive.ics.uci.edu/dataset/45/heart+disease}}~\cite{heart_disease_45}:  
    This dataset, sourced from the UCI Machine Learning Repository (Cleveland subset), contains 297 samples and 13 clinical features. The task is a binary classification of the presence of heart disease.

    \item \textbf{Student Performance (UCI)}\footnote{\url{https://archive.ics.uci.edu/dataset/320/student+performance}}~\cite{student_performance_320}:  
    We use the math-related subset of this dataset (excluding Portuguese-language data), which contains 895 student records with 30 features. The target is binary classification of final-year performance: success ($G3 \geq 10$) vs.\ failure ($G3 < 10$), where $G3$ denotes the final grade.

    \item \textbf{METABRIC}\footnote{\url{https://www.kaggle.com/datasets/raghadalharbi/breast-cancer-gene-expression-profiles-metabric}}~\cite{pereira2016somatic, curtis2012genomic}:  
    A breast cancer gene-expression dataset with 1,904 tumor samples and 489 gene features. Following~\cite{imrie2022composite}, we use progesterone receptor (PR) status as the binary target label.

    \item \textbf{Lung500}:  
    This dataset contains gene-expression profiles for 56 lung cancer patients. Following~\cite{lee2010biclustering}, we select the 500 highest-variance genes from the original 12,625. The task involves four classes: Normal, Carcinoid, Colon Metastasis, and Small Cell Carcinoma.

    \item \textbf{NMNIST}\footnote{\url{https://www-labs.iro.umontreal.ca/~lisa/icml2007data/}}~\cite{larochelle2007empirical, lecun2002gradient}:  
    A variant of MNIST, also referred to as mnist-back-rand, in which each image is overlaid with uniformly sampled background noise. The dataset includes 12,000 grayscale images labeled across 10 digit classes.

    \item \textbf{NMNIST 3-8}:  
    A binary subset of the NMNIST dataset containing 1,000 samples (500 each of digits 3 and 8), resulting in a binary classification task.
\end{itemize}

\section{Implementation Details}
\subsection{GroupFS Hyperparameters}
\label{app:hparam_groupFS}

To facilitate reproducibility, we report all hyperparameters used in our study.  
In all GroupFS experiments, the loss weight $\lambda_1 \in \{0.1, 1, 10, 100\}$ is chosen such that $\mathcal{L}_s$ and $\mathcal{L}_f$ have comparable magnitudes during the first training epoch.  
The regularization weight $\lambda_2$ is selected via a coarse grid search, ranging from high values (where all gates remain closed) to low values (where all gates are open). For the model selected based on the lowest loss, most datasets yield a sparse, non-degenerate solution.  
We set $\beta = 1/\lambda_1$ so that the orthogonality term has an effective coefficient of 1.  
The number of groups $C$ is determined either using a heuristic (App.~\ref{app:choose_C_appendix}) or via a small grid search.

We use $K{=}7$ in the self-tuning kernel~\cite{zelnik2004self}, diffusion time $t{=}2$~\cite{lindenbaum2021differentiable}, and STG noise $\sigma{=}0.5$~\cite{yamada2020feature}.  
All models are trained using Adam~\cite{kingma2014adam} with a learning rate of $\text{lr}{=}10^{-3}$ and PyTorch default settings.  
Final models are selected based on the lowest total loss.

The temperature schedule for each Gumbel-Softmax grouping layer is defined as:
\begin{align*}
\text{temp}(e) = \max\big(&\text{min\_t}, 
&\text{start\_t} - (\text{start\_t} - \text{min\_t}) \cdot \tfrac{e}{E} \big),
\end{align*}
where $e$ and $E$ denote the current and total number of epochs, respectively.  
We set the initial temperature at the first epoch to $\text{start\_t} = 10$, and the minimum temperature to $\text{min\_t} = 10^{-2}$ for all runs, allowing soft, exploratory group assignments early in training that gradually become more discrete over time.

\paragraph{Synthetic data (two moons).}

Unless otherwise noted, we train for 500 epochs with a batch size of 100.  
The number of groups $C$ is selected using the heuristic described in Appendix~\ref{app:choose_C_appendix}.  
\begin{itemize}
    \item Table~\ref{tab:moons_ablations} lists the chosen group counts $C$, loss weights $\lambda_1$ and $\lambda_2$, and the search ranges for $\lambda_2$, including the number of uniformly spaced steps.  
    These settings correspond to the correlation and noise ablation experiments shown in Fig.~\ref{fig:synthetic}C and Fig.~\ref{fig:synthetic}D.
    
    \item For the $(C, d)$ grid sweep shown in Fig.~\ref{fig:grid_2moons}, we perform a sweep over approximately 40 values of $\lambda_2$, spanning the full gating spectrum, from all gates closed to all gates open, centered roughly around the best-performing value $\lambda_2 \pm 2$.  
    The selected $(\lambda_1, \lambda_2)$ values that yield the lowest loss for each grid point are reported in Table~\ref{tab:moons_C_d_grid}.
\end{itemize}

\paragraph{Real-world datasets.}
For each real-world dataset, the group count $C$ is chosen as a local minimum of the heuristic described in App.~\ref{app:choose_C_appendix}.  
The selected values for $C$, batch size (BS), number of training epochs, loss weights $\lambda_1$ and $\lambda_2$, and the search ranges for $\lambda_2$ are summarized in Table~\ref{tab:groupfs_hparams_real_data},  
reported separately for the fixed-budget scenario (Table~\ref{tab:fixed_budget}) and the adaptive-budget scenario (Table~\ref{tab:changed_budget_main}).

\begin{table}[t]
\centering
\scriptsize
\setlength{\tabcolsep}{4pt}

\begin{tabular}{|c|c|c|c|c|c|}
\hline
\textbf{Sweep} & \textbf{Value} & \textbf{C} &
$\boldsymbol{\lambda_1}$ & $\boldsymbol{\lambda_2}$ &
$\lambda_2$ \textbf{range (steps)} \\ \hline
\multicolumn{6}{|c|}{Varying Noise STD} \\ \hline
$\sigma$ & 0.00 & 12 & 1 & 6.3 & $[3.5,\,8]$ (45) \\
$\sigma$ & 0.05 & 12 & 1 & 6.2 & $[3.5,\,8]$ (45) \\
$\sigma$ & 0.10 & 12 & 1 & 6.0 & $[3.5,\,8]$ (45) \\
$\sigma$ & 0.15 & 12 & 1 & 5.8 & $[3.5,\,8]$ (45) \\
$\sigma$ & 0.20 & 12 & 1 & 5.6 & $[3.5,\,8]$ (45) \\
$\sigma$ & 0.25 & 12 & 1 & 5.4 & $[3.5,\,8]$ (45) \\
$\sigma$ & 0.30 & 12 & 1 & 5.2 & $[3.5,\,8]$ (45) \\
$\sigma$ & 0.35 & 12 & 1 & 5.0 & $[3.5,\,8]$ (45) \\
$\sigma$ & 0.40 & 12 & 1 & 4.8 & $[3.5,\,8]$ (45) \\
$\sigma$ & 0.45 & 12 & 1 & 4.7 & $[3.5,\,8]$ (45) \\ \hline
\multicolumn{6}{|c|}{Varying  $\rho$} \\ \hline
$\rho$ & 0.60 &  3 & 1 & 0.6 & $[0,\,2]$  (20) \\
$\rho$ & 0.65 &  3 & 1 & 0.7 & $[0,\,2]$  (20) \\
$\rho$ & 0.70 &  3 & 1 & 0.8 & $[0,\,4]$  (40) \\
$\rho$ & 0.75 &  3 & 1 & 0.9 & $[0,\,4]$  (40) \\
$\rho$ & 0.80 & 12 & 1 & 4.2 & $[2,\,6]$  (40) \\
$\rho$ & 0.85 & 12 & 1 & 4.8 & $[3,\,7]$  (40) \\
$\rho$ & 0.90 & 12 & 1 & 5.5 & $[4,\,8]$  (40) \\
$\rho$ & 0.95 & 12 & 1 & 6.2 & $[5,\,9]$  (40) \\
$\rho$ & 1.00 & 12 & 1 & 7.0 & $[5,\,10]$ (50) \\ \hline
\end{tabular}

\caption{GroupFS hyperparameters for Two-Moons: varying additive noise (top block) and inter-group correlation strength $\rho$ (bottom block).}

\label{tab:moons_ablations}
\end{table}

\begin{table*}[t]
\centering
\scriptsize
\setlength{\tabcolsep}{2pt}
\renewcommand{\arraystretch}{0.9}

\begin{tabular}{|c|cccccccc|}
\hline
\textbf{C} & $d{=}17$ & $d{=}18$ & $d{=}19$ & $d{=}20$ &
            $d{=}21$ & $d{=}22$ & $d{=}23$ & $d{=}24$ \\ \hline
 2  & 1/1.7  & 1/1.4  & 1/1.5  & 1/1.0  & 1/0.9  & 1/1.3  & 1/1.1  & 1/0.7  \\
 3  & 10/2.2 & 10/1.8 & 1/1.8  & 1/1.5  & 1/1.3  & 1/1.1  & 1/1.0  & 1/0.9  \\
 4  & 10/3.0 & 10/2.6 & 1/2.4  & 1/2.1  & 1/1.6  & 1/1.4  & 1/1.4  & 1/1.3  \\
 5  & 10/3.7 & 10/3.2 & 1/2.7  & 1/2.3  & 1/2.2  & 1/1.8  & 1/1.7  & 1/1.5  \\
 6  & 10/4.7 & 10/3.9 & 1/3.5  & 1/2.9  & 1/2.7  & 1/2.1  & 1/2.1  & 1/1.9  \\
 7  & 10/5.3 & 10/4.8 & 1/4.0  & 1/3.6  & 1/3.3  & 1/2.6  & 1/2.5  & 1/2.3  \\
 8  & 10/5.8 & 10/5.2 & 1/4.5  & 1/3.9  & 1/3.5  & 1/3.2  & 1/2.8  & 1/2.5  \\
 9  & 10/6.8 & 10/5.8 & 1/5.4  & 1/4.6  & 1/4.0  & 1/3.6  & 1/3.1  & 1/2.9  \\
10  & 10/7.4 & 10/6.8 & 1/5.8  & 1/5.2  & 1/4.5  & 1/3.9  & 1/3.6  & 1/3.2  \\
11  & 10/8.6 & 10/7.3 & 1/6.6  & 1/5.7  & 1/4.9  & 1/4.5  & 1/4.0  & 1/3.5  \\
12  & 10/10.3& 10/8.3 & 1/7.2  & 1/6.2  & 1/5.4  & 1/4.9  & 1/4.3  & 1/3.8  \\
13  & 10/10.0& 10/9.1 & 1/7.4  & 1/6.8  & 1/5.9  & 1/5.2  & 1/4.6  & 1/4.1  \\
14  & 10/12.0& 10/9.3 & 1/8.4  & 1/7.1  & 1/6.4  & 1/5.5  & 1/4.9  & 1/4.4  \\
15  & 10/12.0& 10/9.6 & 1/8.4  & 1/7.3  & 1/6.7  & 1/5.9  & 1/5.3  & 1/4.8  \\
16  & 10/11.3& 10/9.7 & 1/8.9  & 1/7.8  & 1/6.7  & 1/5.9  & 1/5.4  & 1/5.0  \\
17  & 10/13.0& 10/12.0& 1/10.2 & 1/8.9  & 1/7.6  & 1/6.8  & 1/6.0  & 1/5.3  \\
18  & --      & 10/14.0& 1/10.9 & 1/9.3  & 1/8.0  & 1/7.1  & 1/6.4  & 1/5.7  \\
19  & --      & --      & 1/10.8 & 1/9.3  & 1/8.4  & 1/7.3  & 1/6.5  & 1/6.0  \\
20  & --      & --      & --      & 1/9.6  & 1/9.0  & 1/7.7  & 1/6.8  & 1/6.3  \\
21  & --      & --      & --      & --      & 1/8.7  & 1/8.1  & 1/7.1  & 1/6.4  \\
22  & --      & --      & --      & --      & --      & 1/8.1  & 1/7.5  & 1/6.6  \\
23  & --      & --      & --      & --      & --      & --      & 1/7.4  & 1/7.0  \\
24  & --      & --      & --      & --      & --      & --      & --      & 1/6.9  \\ \hline
\end{tabular}

\caption{
GroupFS hyperparameters for Two-Moons: varying feature dimension $d$ and group count $C$.  
Each cell reports the selected $\lambda_1 / \lambda_2$.}

\label{tab:moons_C_d_grid}
\end{table*}

\begin{table*}[t]
\centering
\scriptsize
\setlength{\tabcolsep}{3pt}
\renewcommand{\arraystretch}{0.9}

\begin{tabular}{|l|cccccc|cccccc|}
\hline
\multirow{2}{*}{\textbf{Dataset}} &
\multicolumn{6}{c|}{Fixed-budget} &
\multicolumn{6}{c|}{Adaptive-budget} \\ \cline{2-13}
 & \textbf{Epochs} & \textbf{BS} & \textbf{C} & $\boldsymbol{\lambda_1}$ & $\boldsymbol{\lambda_2}$ & $\boldsymbol{\lambda_2}$ \textbf{Range (steps)} &
   \textbf{Epochs} & \textbf{BS} & \textbf{C} & $\boldsymbol{\lambda_1}$ & $\boldsymbol{\lambda_2}$ & $\boldsymbol{\lambda_2}$ \textbf{Range (steps)} \\ \hline
ALLAML        & 4000 &  50 & 26 & 10   & 1.45 & 1-1.6 (12)     & 4000 &  50 & 26 & 10   & 1.30 & 1-1.6 (12) \\
Lung500       & 5000 &  32 & 22 & 10   & 20.5 & 22-25 (30)     & 5000 &  32 & 22 & 10   & 23.9 & 22-25 (30) \\
METABRIC      & 1000 &  100 & 16 & 1    & 0.67 & 0.5-1 (50)     & 1000 &  100 & 16 & 1    & 0.62 & 0.5-1 (50) \\
HeartDisease  & 1000 &  100 &  6 & 1    & 1.87 & 1.5-1.95 (45)     & 1000 &  100 &  6 & 1.5    & 1.73 & 1.5-1.95 (45) \\
Yale          & 2000 &  32 & 16 & 10   & 2.89 & 2.5-3.5 (100)  & 2000 &  32 & 16 & 10   & 3.17 & 2.5-3.5 (100) \\
AR10P         & 2000 &  32 & 16 & 10   & 2.89 & 6.7-8 (30)     & 2000 &  32 & 16 & 10   & 3.17 & 6.7-8 (30) \\
PIE10P        & 1000 &  50 & 20 & 100  & 7.55 & 25–35 (20)     & 1000 &  50 & 20 & 100  & 7.90 & 25–35 (20) \\
MNIST 3-8     & 1000 & 100 & 29 & 100  & 34.0 & 0.05-0.45 (20) & 1000 & 100 & 29 & 100  & 33.5 & 0.05-0.45 (20) \\
MNIST         & 1000 & 100 &  7 & 1    & 0.23 & 0.1-0.13 (30)  & 1000 & 100 &  7 & 1    & 0.15 & 0.1-0.13 (30) \\
Student       & 1000 &  70 &  7 & 0.1  & 0.66 & 0.5-0.7 (20)              & —    &  —  & —  & —    & —    & — \\ \hline
\end{tabular}

\caption{
GroupFS hyperparameters for real-world datasets.}
\label{tab:groupfs_hparams_real_data}
\end{table*}

\subsection{Baseline Hyperparameters}
\label{app:hparam_baselines}
We trained all baseline methods using the official code released by the authors of the respective method papers, with default settings unless stated otherwise.
\begin{itemize}
    \item \textbf{LS}~\cite{he2005laplacian} and \textbf{MCFS}~\cite{cai2010unsupervised}:  
      We use the scikit-feature implementations\footnote{\url{https://github.com/jundongl/scikit-feature}}~\cite{li2018feature}, 
      with the self-tuning kernel from eq.~(\ref{eq:w_self}) ($K=7$ neighbors), these methods require no additional hyperparameter tuning.

    \item \textbf{CAE}~\cite{abid2019concrete}:  
          Experiments are based on the authors’ Python code\footnote{\url{https://github.com/mfbalin/Concrete-Autoencoders}}.  
         We conduct a grid search over the following settings:
          \begin{itemize}
              \item Batch size: $\{32, 64, 128\}$ for small datasets; $\{64, 128, 256, 512\}$ for large ones.
              \item Hidden widths: $\left\lfloor\frac{4k}{9}\right\rfloor$, $\left\lfloor\frac{2k}{3}\right\rfloor$, $\left\lfloor\frac{3k}{2}\right\rfloor$, with $k$ being the number of selected features, as in~\cite{abid2019concrete}.
          \end{itemize}
          Each configuration is trained for 200 epochs using a single trial.  
          We follow the original setup: Adam optimizer, learning rate $10^{-3}$, initial temperature 10, minimum temperature $2 \times 10^{-3}$, and exponential temperature annealing.  

          \emph{Fixed-budget setting}: a 90/10 train-validation split; we report the model with the lowest reconstruction loss on the validation set.  
          \emph{Adaptive-budget setting}: trained on the full dataset; we select the checkpoint with the highest $k$-means accuracy (based on selected features). Table~\ref{tab:cae_hparams} summarizes the selected batch size (BS) and hidden-layer width (HW) for both settings across all datasets.

\begin{table}[t]
\centering
\scriptsize
\setlength{\tabcolsep}{5pt}

\begin{tabular}{|l|cc|cc|}
\hline
\multirow{2}{*}{\textbf{Dataset}} &
\multicolumn{2}{c|}{Fixed-budget} &
\multicolumn{2}{c|}{Adaptive-budget} \\ \cline{2-5}
 & \textbf{BS} & \textbf{HW} & \textbf{BS} & \textbf{HW} \\ \hline
ALLAML          & 64  & 411 & 64  & 150 \\
Lung500         & 32  & 351 & 16  & 177 \\
METABRIC        & 512 & 150 & 256 &  75 \\
HeartDisease    & 20  &   4 & 20  &   3 \\
Yale            & 32  & 151 & 128 & 177 \\
AR10P           & 32  & 543 & 128 & 150 \\
PIE10P          & 64  &  21 & 128 & 177 \\
NMNIST 3-8      & 256 &  76 &  64 & 600 \\
NMNIST          & 64  &  81 &  64 & 600 \\
Student         & 20  &  13 & —   & —   \\ \hline
\end{tabular}

\caption{CAE hyperparameters for real-world datasets.}
\label{tab:cae_hparams}
\end{table}

    \item \textbf{DUFS}~\cite{lindenbaum2021differentiable}:  
          We use the PyTorch implementation from featselectlib\footnote{\url{https://github.com/LindenbaumLab/project-featselectlib}}.  
          We tune the regularization strength $\lambda \in \{10^{-3}, 10^{-2}, 10^{-1}, 1\}$ and learning rate $\in \{10^{-4}, 10^{-3}, 10^{-2}, 10^{-1}\}$ using Adam.  
          Batch size and number of epochs follow Table~\ref{tab:groupfs_hparams_real_data}. Table~\ref{tab:dufs_hparams} reports the selected $\lambda$ and learning rate (LR) for both settings across all datasets.

\begin{table}[t]
\centering
\scriptsize
\setlength{\tabcolsep}{5pt}

\begin{tabular}{|l|cc|cc|}
\hline
\multirow{2}{*}{\textbf{Dataset}} &
\multicolumn{2}{c|}{Fixed-budget} &
\multicolumn{2}{c|}{Adaptive-budget} \\ \cline{2-5}
 & $\boldsymbol{\lambda}$ & \textbf{LR} & $\boldsymbol{\lambda}$ & \textbf{LR} \\ \hline
ALLAML          & 0.01  & 0.01   & 0.01  & 0.0001 \\
Lung500         & 0.001 & 0.01   & 0.1   & 0.1    \\
METABRIC        & 0.001 & 0.01   & 0.01  & 0.1    \\
HeartDisease    & 0.001 & 0.1   & 1  & 0.0001 \\
Yale            & 0.001 & 0.01   & 0.1   & 0.1    \\
AR10P           & 0.001 & 0.01   & 0.1   & 0.001  \\
PIE10P          & 0.001 & 0.01   & 1     & 0.0001 \\
NMNIST 3-8      & 0.001 & 0.01   & 0.01  & 0.1    \\
NMNIST          & 0.001 & 0.01   & 0.001 & 0.0001 \\
Student         & 0.001 & 0.01   & —     & —      \\ \hline
\end{tabular}

\caption{DUFS hyperparameters for real-world datasets.}
\label{tab:dufs_hparams}
\end{table}

\begin{table}[t]
\centering
\scriptsize
\setlength{\tabcolsep}{4pt}

\begin{tabular}{|l|cccc|cccc|}
\hline
\multirow{2}{*}{\textbf{Dataset}} &
\multicolumn{4}{c|}{Fixed-budget} &
\multicolumn{4}{c|}{Adaptive-budget} \\ \cline{2-9}
 & $\boldsymbol{\alpha}$ & $\boldsymbol{\gamma}$ & $\boldsymbol{\sigma}$ & \textbf{C} &
   $\boldsymbol{\alpha}$ & $\boldsymbol{\gamma}$ & $\boldsymbol{\sigma}$ & \textbf{C} \\ \hline
ALLAML          & 0.1   & 0.1   & 1000  & 10  & 0.1   & 0.1   &  100  &  2 \\
Lung500         & 0.1   & 0.1   &  100  & 20  & 100   & 100   & 1000  & 30 \\
METABRIC        & 0.1   & 0.1   &  100  & 10  & 1     & 1     & 1000  & 20 \\
HeartDisease    & 0.1   & 0.1   &  100  &  2  & 1   & 100 &  10000  &  2 \\
Yale            & 0.1   & 0.1   & 1000  &  2  & 100   & 1     &  100  &  2 \\
AR10P           & 0.1   & 0.1   & 1000  &  2  & 0.1   & 100   & 10000 &  2 \\
PIE10P          & 0.1   & 0.1   & 1000  &  2  & 0.1   & 100   & 1000  & 30 \\
NMNIST 3-8      & 0.1   & 0.1   &  100  & 10  & 1     & 0.1   & 1000  & 30 \\
NMNIST          & —     & —     & —     & —   & —     & —     & —     & —  \\
Student         & 0.1   & 0.1   &  100  &  2  & —     & —     & —     & —  \\ \hline
\end{tabular}

\caption{MGAGR hyperparameters for real-world datasets.}
\label{tab:mgagr_hparams}
\end{table}

    \item \textbf{MGAGR}~\cite{you2023mgagr}:
          We use the MATLAB implementation\footnote{\url{https://github.com/misteru/MGAGR}} with a subset of the authors’ recommended grid:  
            regularization strengths $\alpha \in \{0.1, 1, 10\}$ and $\gamma \in \{0.1, 1, 10\}$,  
            kernel size $\sigma \in \{10, 100, 1000\}$,  
            and number of feature groups $C \in \{2, 10, 20, 30\}$ (limited by feature dimensionality).

          Grouping follows the "average grouping" strategy, identified as best in the original paper, with a maximum of 10 iterations. Table~\ref{tab:mgagr_hparams} summarizes the selected $\alpha$, $\gamma$, $\sigma$, and $C$ values for both settings across all datasets.

            We attempted to evaluate MGAGR on the NMNIST dataset (12,000 samples, 784 features) using the official MATLAB implementation. However, due to the algorithm's nested-loop design, the runtime was directly affected by the sample size. Each hyperparameter setting required approximately 9.5 hours per iteration, and running 10 iterations on an Intel Xeon Gold 6230 CPU would take around 95 hours. A full sweep of 108 hyperparameter configurations would thus exceed 10,200 CPU-hours. As such, we omit MGAGR results for this dataset due to impractical runtime. Notably, the original MGAGR paper~\cite{you2023mgagr} did not report results on datasets of this scale.

    \item \textbf{CompFS (supervised)}~\cite{imrie2022composite}: 
          We adapt the official PyTorch implementation\footnote{\url{https://github.com/a-norcliffe/Composite-Feature-Selection}} to use a 90/10 train-validation split.  
          Grid search is conducted over:
          \begin{itemize}
              \item Batch size: $\{50, 100\}$ for small datasets; $\{50, 100, 500\}$ for large ones.
              \item Hidden dimension: $\{128, 256, 512\}$, omitting values that exceed the input dimension.  
                    For Heart Disease and Student Performance, we use $\{4, 8, 10\}$.
              \item Loss weights:  
                    \begin{itemize}
                        \item Fixed-budget: $\beta_s, \beta_d \in \{0.18, 0.6, 1.2, 3, 4.5, 6\}$  
                        \item Adaptive-budget: $\beta_s, \beta_d \in \{0.18, 1.2, 4.5, 6\}$
                    \end{itemize}
              \item Number of learners: we use powers-of-two divisors of $C$ in Table~\ref{tab:groupfs_hparams_real_data}, i.e., $C/2$, $C/4$, $C/8$, and so on, stopping when the number falls below 1.
          \end{itemize}
          Training uses Adam with a fixed learning rate of $10^{-3}$.  
            We retain the model with the highest validation accuracy under the supervised loss. However, the values reported in Tables~\ref{tab:changed_budget_main} and~\ref{tab:fixed_budget} reflect $k$-means clustering accuracy computed on the features selected by that model, rather than the accuracy of the supervised classifier itself.

            Table~\ref{tab:compfs_hparams} summarizes the selected batch size (BS), hidden dimension ($h_{\text{dim}}$), loss-weight pair ($\beta_d$, $\beta_s$) , and number of learners ($N_L$)  for both settings across all datasets.

\begin{table*}[t]
\centering
\scriptsize
\setlength{\tabcolsep}{4pt}

\begin{tabular}{|l|ccccc|ccccc|}
\hline
\multirow{2}{*}{\textbf{Dataset}} &
\multicolumn{5}{c|}{Fixed-budget} &
\multicolumn{5}{c|}{Adaptive-budget} \\ \cline{2-11}
 & \textbf{BS} & $\boldsymbol{h_{\!dim}}$ & $\boldsymbol{\beta_d}$ & $\boldsymbol{\beta_s}$ & $\boldsymbol{N_L}$ &
   \textbf{BS} & $\boldsymbol{h_{\!dim}}$ & $\boldsymbol{\beta_d}$ & $\boldsymbol{\beta_s}$ & $\boldsymbol{N_L}$ \\ \hline
ALLAML          & 30  & 128 & 0.18 & 0.18 &  6 & 30  & 128 & 6    & 6    & 26 \\
Lung500         & 30  & 128 & 0.18 & 0.18 &  5 & 30  & 128 & 4.5  & 4.5  & 22 \\
METABRIC        & 100 & 512 & 0.18 & 0.18 &  8 & 50  & 128 & 0.18 & 0.18 &  5 \\
HeartDisease    & 50  &  8 & 1.2 & 0.60 &  3 & 50  &   8 & 6    & 6    &  6 \\
Yale            & 50  & 128 & 4.50 & 6.00 &  4 & 50  & 256 & 1.2  & 6    &  6 \\
AR10P           & 50  & 128 & 0.18 & 0.18 & 10 & 50  & 512 & 4.5  & 4.5  &  8 \\
PIE10P          & 50  & 512 & 1.20 & 0.18 & 15 & 100 & 128 & 0.18 & 4.5  & 15 \\
NMNIST 3–8      & 100 & 256 & 0.18 & 4.50 &  7 & 100 & 128 & 4.5  & 6    &  7 \\
NMNIST          & 50   & 256   & 0.18    & 4.5    & 4  & 50  & 128 & 4.5  & 0.18 & 17 \\
Student         & 50  &   8 & 0.18 & 0.60 &  7 & —   & —   & —    & —    & —  \\ \hline
\end{tabular}

\caption{CompFS hyperparameters for real-world datasets.}
\label{tab:compfs_hparams}
\end{table*}
\end{itemize}

\subsection{Computing Infrastructure}
Our method was run on either a local machine or GPU nodes on a SLURM-managed cluster. The local system includes a 13th Gen Intel Core i7-13700 CPU (2.10\,GHz), 64\,GB RAM, and an NVIDIA RTX A2000 GPU (12\,GB VRAM), running Windows 11 with CUDA 12.8. SLURM jobs used NVIDIA L40S GPUs (46\,GB VRAM), also with CUDA 12.8.

All baseline methods, except MGAGR, were executed on the same local machine. MGAGR was run on a CPU-only cluster using Intel Xeon Gold 6230 CPUs (2.10\,GHz). All experiments used Python 3.10, with key libraries including PyTorch 2.5.1.
To ensure reproducibility, we initialized the random seed at the start of each run.

\subsection{Computational Complexity}
For GroupFS, the computational complexity is governed by the sample size $N$ and the number of features $d$. The number of groups $C$ is relatively small (a few tens) in most practical settings. Initialization includes constructing a $d \times d$ feature graph using all $N$ samples, which leads to $\mathcal{O}(Nd^2)$ operations, and computing its eigendecomposition, leads to up to $\mathcal{O}(d^3)$. During training, in the full-batch case ($B = N$), the complexity of updating the sample graph on each epoch is $\mathcal{O}(N^2 d)$. For $t > 1$ diffusion steps, explicitly multiplying $N \times N$ matrices $t$ times adds $\mathcal{O}(tN^3)$ time per epoch.

\section{Sign Conventions in the Loss}
\label{app:signs}
The sample-wise smoothness loss term \(\mathcal{L}_{s}\) is based on the random-walk matrix  
\(P = D^{-1} W\)~\cite{spielman2025sagt}.  
Since \(P\) is generally non-symmetric, its eigenvectors are not orthogonal, and the smooth (i.e., low-frequency) modes correspond to the largest eigenvalues.  
As a result, \(\mathcal{L}_{s}\) must be maximized, and we therefore include it in the objective with a negative sign.

In contrast, the feature-wise smoothness loss term \(\mathcal{L}_{f}\) is based on the symmetric normalized Laplacian  
\(L_{\text{sym}} = I - D^{-1/2} W D^{-1/2}\)~\cite{ng2001spectral, von2007tutorial}.  
Here, the eigenvectors form an orthonormal basis, and the smooth modes correspond to the smallest eigenvalues.  
Thus, minimizing \(\mathcal{L}_{f}\) naturally promotes smoothness, and we include this term with a positive sign.

To prevent the learned feature representation \(\mathbf{F}\) from collapsing into degenerate directions, we add an orthonormality regularization term:
\[
\left\|\mathbf{F}^{\top} \mathbf{F} - I \right\|_{F}^{2}.
\]

\begin{figure*}[t]
  \centering
 \includegraphics[width=\textwidth]{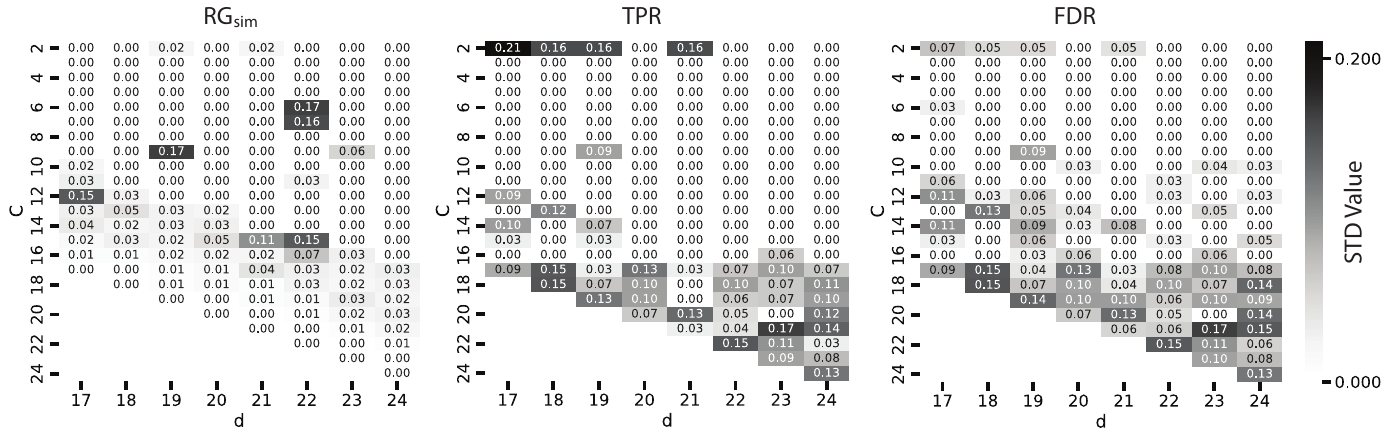}

  \caption{\textbf{Two-moons run-to-run variability: Effect of feature dimension $d$ and group count $C$.}
    Standard deviation of $RG_{\text{sim}}$, TPR, and FDR for the best-loss model over 10 random seeds.}\label{fig:grid_2moons_std}
\end{figure*}

\section{Selecting the Number of Clusters $\boldsymbol{C}$}
\label{app:choose_C_appendix}
To determine the number of clusters, we follow the approach of \citet{zelnik2004self}.  
After computing the spectral embedding, we apply the closed-form Procrustes alignment of \citet{schonemann1966generalized} to match the (arbitrarily rotated) embedding to a binary cluster-indicator matrix.  
For each candidate $C \in \{2,3,\dots,C_{\max}\}$, we compute the distortion score $\mathcal{E}(C)$ and choose the value of $C$ that minimizes it or corresponds to a local minimum. 

\begin{enumerate}
    \item Construct the normalized graph Laplacian (See Sec.\ref{sec:graphs}) of the feature graph, $L_{\mathrm{feat}} \in \mathbb{R}^{d\times d}$:  
    \[
    L_{\mathrm{feat}} = I - D^{-1/2} W D^{-1/2}
    \]
    Extract its \( C \) smallest eigenvectors, stack them column-wise to form  
    \( U_C \in \mathbb{R}^{d \times C} \), and normalize each row:
    \[
    \widetilde{U}_C(i,:) = \frac{U_C(i,:)}{\lVert U_C(i,:) \rVert_2}
    \]
    as recommended by \citet{von2007tutorial}.

    \item Run \( k \)-means with \( k = C \) on the rows of \( \widetilde{U}_C \), producing cluster labels  
    \( \ell \in \{1, \dots, C\}\) for each row.

    \item Form the binary cluster indicator matrix  
    \( Y \in \{0,1\}^{d \times C} \), where \( Y_{ij} = \mathbb{1}\{\ell_i = j\} \), and \( \mathbb{1} \) is the indicator function.

    \item Solve the orthogonal Procrustes problem:
    \[
    R^{\star} = \argmin_{R^\top R = I} \left\| \widetilde{U}_C R - Y \right\|_F^2
    \]
    where
    \[
    \widetilde{U}_C^\top Y = U \Sigma V^\top, \quad R^\star = U V^\top
    \]

    \item Compute the distortion score:
    \[
    \mathcal{E}(C) = \left\| \widetilde{U}_C R^\star - Y \right\|_F^2
    \]
      the Frobenius error between the rotated embedding and the
      one-hot cluster-indicator matrix~\(Y\); lower values indicate that the features can be cleanly partitioned into \(C\) groups
\end{enumerate}

\section{Additional Experimental Results}
\label{app:extra_results}

This section presents supplementary experimental findings that were omitted from the main paper due to space constraints.

\subsection{Synthetic Data.}
\label{app:exp_synthetic}
As a complementary analysis to Figure~\ref{fig:grid_2moons}, we report in Figure~\ref{fig:grid_2moons_std} the standard deviation across 10 runs with different random seeds for each combination of group count $C$ and feature dimensionality $d$. For each $(C, d)$ configuration, we used the $(\lambda_1, \lambda_2)$ pair that achieved the lowest total loss (Table~\ref{tab:moons_C_d_grid}).

We observe that the standard deviation of $RG_{\text{sim}}$ is generally low, with only a few non-zero values, indicating consistent group discovery across seeds despite stochastic training dynamics.  
For TPR and FDR, we find that when $C \leq 2 + (d - 10)$ (excluding the special case $C=2$), the standard deviations are small or zero, suggesting stable selection behavior.  
In contrast, when $C > 2 + (d - 10)$, variability increases, particularly in TPR and FDR, indicating that excessive group capacity can lead to inconsistent feature selection.

\subsection{Real-World Data.}
\label{app:exp_real_data}

\begin{table*}[t]
\centering
\setlength{\tabcolsep}{0.7mm}                   
\begin{small}                                       
\begin{tabular*}{\textwidth}{@{\extracolsep{\fill}}lccccccccc}
\toprule
Dataset & ALL & LS & MCFS & CAE & DUFS & MGAGR & CompFS & GroupFS & \#Feat \\
\midrule
ALLAML            & 9.5$\pm$10.1  & \textbf{15.7$\pm$2.4} & 14.1$\pm$7.9  & 10.0$\pm$6.2  & 8.2$\pm$9.0  & 6.7$\pm$7.0  & 0.5$\pm$6.1  & 15.5$\pm$2.4  & 274 \\
Lung500      & 82.4$\pm$11.8 & 77.4$\pm$5.4 & 76.7$\pm$9.6 &  85.4$\pm$7.1&  82.2$\pm$8.6 & 76.2$\pm$10.0& 76.1$\pm$10.0 & \textbf{88.7$\pm$8.3}  & 234 \\
METABRIC          & 11.4$\pm$7.6  & 9.0$\pm$6.4   & 10.6$\pm$7.0  & 16.6$\pm$0.2  & 6.9$\pm$8.7  & \textbf{18.2$\pm$6.1} & 8.2$\pm$5.5  & 13.3$\pm$4.8  & 226 \\
Heart Disease     &  42.1$\pm$1.9 &  40.5$\pm$3.7 &   42.3$\pm$1.0& 32.6$\pm$2.6 & 31.8$\pm$10.8 & 274.0$\pm$9.1  &  40.7$\pm$1.1 &  \textbf{43.6$\pm$1.3} & 10 \\
Yale              & \textbf{27.8$\pm$3.3} & 23.1$\pm$1.9  & 18.5$\pm$3.0  & 27.3$\pm$3.2  & 23.4$\pm$4.6  & 18.8$\pm$3.9 & 25.4$\pm$5.9  & 22.5$\pm$2.0  & 341 \\
AR10P             & 5.0$\pm$3.0   & 4.9$\pm$0.7   & 2.9$\pm$2.1   & 0.8$\pm$1.2   & 4.1$\pm$1.4   & 4.2$\pm$2.4   & 4.6$\pm$2.5   & \textbf{10.2$\pm$3.0}  & 362 \\
PIE10P            & 9.0$\pm$2.1   & 3.0$\pm$0.7   & \textbf{15.7$\pm$2.3} & 5.6$\pm$1.3   & 15.2$\pm$2.7  & 11.1$\pm$2.3  & 8.9$\pm$2.5   & 13.1$\pm$2.1  & 49 \\
NMNIST 3-8    & 22.6$\pm$9.1  & 29.8$\pm$2.2  & 11.8$\pm$0.3  & 1.9$\pm$0.2   & 0.0$\pm$0.1   & 0.2$\pm$0.2   & 8.5$\pm$1.1   & \textbf{44.2$\pm$0.3}  & 51 \\
NMNIST       & \textbf{31.4$\pm$1.9} & 27.2$\pm$1.3  & 29.8$\pm$1.0  & 26.4$\pm$1.7  & 3.4$\pm$0.2   & --             & 27.0$\pm$1.0  & 30.2$\pm$2.4  & 184 \\
\bottomrule
\end{tabular*}
\end{small}

\caption{\textbf{Scenario 1 - ARI - fixed budget, unsupervised setting.} 
Mean $\pm$ std.\ Adjusted Rand Index \textbf{(\%)} over 10 runs.  
All methods use the same feature budget (\#Feat). Bold marks the best score per dataset.}
\label{tab:ari_comparison}
\end{table*}

\begin{table*}[t]
\centering
\setlength{\tabcolsep}{1mm}
\begin{small}
\begin{tabular}{lccccccc}
\toprule
Dataset & LS & MCFS & CAE & DUFS & MGAGR & CompFS & GroupFS \\
\midrule
ALLAML        & 72.2${}\pm{}$0.0 & 71.8${}\pm{}$1.3 & 70.4${}\pm{}$6.3 & 71.5${}\pm{}$9.3 & 66.3${}\pm{}$7.1 & 67.4${}\pm{}$6.4 & \textbf{72.8${}\pm{}$1.3} \\
Lung500       & 82.2${}\pm{}$4.7 & 87.5${}\pm{}$9.7 & 92.3${}\pm{}$8.2 & 95.0${}\pm{}$6.8 & 93.0${}\pm{}$8.3 & 94.5${}\pm{}$5.6 & \textbf{96.1${}\pm{}$6.4} \\
METABRIC      & 68.0${}\pm{}$5.8 & 69.3${}\pm{}$0.1 & 72.8${}\pm{}$0.2 & 72.4${}\pm{}$0.2 & 71.6${}\pm{}$0.3 & 73.2${}\pm{}$0.1 & \textbf{73.4${}\pm{}$0.1} \\
HeartDisease  & 81.9${}\pm{}$1.3 & 82.7${}\pm{}$0.3 & 82.8${}\pm{}$0.2 & 83.5${}\pm{}$1.0 & \textbf{84.3${}\pm{}$0.2} & \textbf{84.3${}\pm{}$0.7} & 83.1${}\pm{}$0.5 \\
Yale          & 43.0${}\pm{}$3.9 & 45.0${}\pm{}$4.6 & 45.6${}\pm{}$2.0 & 42.6${}\pm{}$3.1 & 40.7${}\pm{}$3.8 & 46.7${}\pm{}$3.5 & \textbf{48.3${}\pm{}$1.3} \\
AR10P         & 32.9${}\pm{}$3.2 & 29.5${}\pm{}$2.2 & 30.1${}\pm{}$3.0 & 34.2${}\pm{}$3.0 & 32.6${}\pm{}$3.5 & 29.8${}\pm{}$3.5 & \textbf{34.7${}\pm{}$3.9} \\
PIE10P        & 26.3${}\pm{}$1.7 & 34.1${}\pm{}$3.0 & 26.6${}\pm{}$1.8 & \textbf{42.1${}\pm{}$2.1} & 36.0${}\pm{}$2.6 & 29.1${}\pm{}$1.6 & 40.8${}\pm{}$3.7 \\
NMNIST 3–8    & 76.6${}\pm{}$0.7 & 68.1${}\pm{}$1.0 & 76.0${}\pm{}$0.5 & 80.2${}\pm{}$0.4 & 77.4${}\pm{}$1.0 & 80.4${}\pm{}$4.8 & \textbf{84.1${}\pm{}$0.5} \\
NMNIST        & 49.4${}\pm{}$3.0 & 50.4${}\pm{}$2.7 & 50.5${}\pm{}$3.1 & 42.0${}\pm{}$1.7 & -- & \textbf{55.0${}\pm{}$3.5} & 48.9${}\pm{}$2.7 \\
\bottomrule
\end{tabular}
\end{small}
\caption{\textbf{Scenario 2 - Adaptive budget, accuracy‑guided setting.} $k$‑means accuracy (mean$\pm$std.\ over 10 runs). Complementary to Table~\ref{tab:changed_budget_main}.}
\label{tab:changed_budget_std}
\end{table*}

\begin{table*}[t]
\centering
\setlength{\tabcolsep}{1mm} 
\begin{small}                
\begin{tabular*}{\textwidth}{@{\extracolsep{\fill}}clp{0.72\textwidth}}
\toprule
\textbf{Idx} & \textbf{Feature} & \textbf{Description} \\
\midrule
0  & school     & Pupil’s school (GP / MS) \\
1  & sex        & Female (F) or male (M) \\
2  & age        & Age (15-22) \\
3  & address    & Home address type (U = urban, R = rural) \\
4  & famsize    & Family size ($\le3$ / $>3$) \\
5  & Pstatus    & Parents’ co-habitation status (T / A) \\
6  & Medu       & Mother’s education (0-4) \\
7  & Fedu       & Father’s education (0-4) \\
8  & Mjob      & Mother’s job (teacher, health, services, at\_home, other) \\
9  & Fjob       & Father’s job (same categories as \texttt{Mjob}) \\
10 & reason     & Reason for school choice (home, reputation, course, other) \\
11 & guardian   & Guardian (mother, father, other) \\
12 & traveltime & Home-school travel time (1 $<$15 min … 4 $>$1 h) \\
13 &studytime  & Weekly study time (1 $<$2 h … 4 $>$10 h) \\
14 & failures   & Past class failures (0-3; 4 = $\ge$4) \\
15 & schoolsup  & Extra educational support (yes/no) \\
16 & famsup     & Family educational support (yes/no) \\
17 & paid      & Extra paid classes (yes/no) \\
18 & activities & Extra-curricular activities (yes/no) \\
19 & nursery    & Attended nursery school (yes/no) \\
20 & higher     & Wants to take higher education (yes/no) \\
21 & internet   & Internet access at home (yes/no) \\
22 & romantic   & In a romantic relationship (yes/no) \\
23 & famrel     & Family relationship quality (1 very bad … 5 excellent) \\
24 & freetime   & Free time after school (1-5) \\
25 & goout      & Going-out frequency (1-5) \\
26 & Dalc       & Work-day alcohol consumption (1-5) \\
27 & Walc       & Weekend alcohol consumption (1-5) \\
28 & health     & Current health status (1-5) \\
29 & absences   & Number of school absences (0-93) \\
\bottomrule
\end{tabular*}
\end{small}

\caption{\textbf{Glossary of Student Performance Features.}
Descriptions of all 30 input features in the UCI Student Performance (Math) dataset.  
Indices correspond to those used in Table~\ref{tab:student_selection}.}
\label{tab:student_features}
\end{table*}

We further evaluate clustering performance using the Adjusted Rand Index (ARI) as an alternative to clustering accuracy.  
Table~\ref{tab:ari_comparison} mirrors the setup of Table~\ref{tab:fixed_budget}, but reports ARI scores computed from $k$-means clustering results. GroupFS achieves the highest ARI on 4 out of 9 datasets.  
Notably, for Yale and NMNIST, the best ARI is obtained by using all features, outperforming all feature selection baselines, consistent with the accuracy-based trends.  
Moreover, no other method consistently ranks among the top performers across datasets, underscoring the robustness of our approach in diverse real-world scenarios.

As a complement to Table~\ref{tab:changed_budget_main}, Table~\ref{tab:changed_budget_std} reports the mean$\pm$standard deviation over 10 runs (with different $k$-means seeds). Due to space constraints, we omit standard deviations in the main table and, conversely, omit the selected feature counts in the appendix table. Both tables summarize the same experiments and provide complementary details.

\subsection{Student Performance (UCI) – Selected Feature Subsets}
\label{app:exp_studentperf}

This section provides a detailed breakdown of feature selection results on the UCI Student Performance dataset, including the specific features selected by each method and the groupings discovered (if applicable). We use the math-course subset of the dataset~\cite{student_performance_320}, as described in Appendix~\ref{app:data_details_real}. Table~\ref{tab:student_features} lists all available features, while Table~\ref{tab:student_selection} presents the subsets selected by each method under a fixed budget of 9 features. GroupFS discovers meaningful groupings. For example, it places alcohol consumption on weekdays and weekends together in one group and the mother’s job with both parents’ education levels in another group.  
In contrast, CompFS produces less meaningful groupings. For instance, it assigns sex, family educational support, weekday alcohol, and health to the same group, despite the lack of a clear conceptual link. 
Furthermore, CompFS assigns single features to separate groups, effectively reverting to individual feature selection.

\begin{table*}[t]
\centering
\setlength{\tabcolsep}{1pt}
\resizebox{\textwidth}{!}{%
\begin{small}
\begin{tabular}{|l|c|c|c|c|c|c|c|}
\hline
\textbf{Feature} & \textbf{LS} & \textbf{MCFS} & \textbf{CAE} & \textbf{DUFS} & \textbf{MGAGR} & \textbf{CompFS} & \textbf{GroupFS}\\
\hline\hline
school      & \tick & \tick &       & \tick &  \tick     &       &            \\ \hline
sex         & \tick &       &       &       &       &    \selg{3}  &            \\ \hline
age         & \tick & \tick & \tick &       &       &            &            \\ \hline
address     &       & \tick &       &       &       &            &    \\ \hline
famsize     &       &       &       &       &       &            &            \\ \hline
Pstatus     &       &       &       & \tick &   \tick    & \selg{1}    &      \\ \hline
Medu        & \tick &       &       &       &       &            &          \selg{3}  \\ \hline
Fedu        & \tick &       & \tick &       &       &            &           \selg{3} \\ \hline
Mjob        & \tick &       &       &       &       &            &           \selg{3} \\ \hline
Fjob        &       &       &       &       &       &            &            \\ \hline
reason      &       &       &       &       &       &            &            \\ \hline
guardian    &       &       & \tick &       &       & \selg{1}    &    \\ \hline
traveltime  &       & \tick &       &       &   \tick   &        &    \\ \hline
studytime   &       & \tick &       &       &       &            &            \\ \hline
failures    &       & \tick & \tick & \tick &   \tick   &            &   \selg{2} \\ \hline
schoolsup   &       &       &       & \tick &   \tick   &            &            \\ \hline
famsup      &       &       &       &       &       & \selg{3}    &            \\ \hline
paid        &       &       & \tick &       &       & \selg{2}    &            \\ \hline
activities  &       &       &       &       &       &            &            \\ \hline
nursery     &       &       &       &       &       &            &            \\ \hline
higher      & \tick & \tick &       & \tick &  \tick   &            &   \selg{2} \\ \hline
internet    &       &       & \tick &       &   \tick   &            &            \\ \hline
romantic    &       &       &       & \tick &       &            &    \selg{2} \\ \hline
famrel      &       &       &       & \tick &       &            &            \\ \hline
freetime    &       &       &       & \tick &       &            &            \\ \hline
goout       &       &       &       &       &       & \selg{5}    &            \\ \hline
Dalc        & \tick & \tick & \tick &       & \tick   & \selg{3}    & \selg{1}   \\ \hline
Walc        & \tick & \tick & \tick &       &       &            & \selg{1}   \\ \hline
health      &       &       &       &       &       & \selg{3}    &            \\ \hline
absences    &       &       &       & \tick &  \tick   & \selg{4}    &   \selg{2} \\ \hline
\textbf{Accuracy (\%)} & 58.6$\pm$3.0 & 65.3$\pm$0.1 & 61.7$\pm$2.7 & 66.7$\pm$3.3 &  65.2$\pm$2.7 & 56.5$\pm$3.0 & 61.3$\pm$2.6 \\ \hline
\end{tabular}
\end{small}
}
\vspace{0.5em}
\caption{\textbf{Feature Selection Comparison on Student-Performance Dataset (budget = 9 features).}  
Each method is limited to selecting at most nine features; however, CAE effectively chooses only eight because two of its selections collapse onto the same feature.  
Each cell shows whether a feature was retained.  
\tick\; marks an individually chosen feature, while \selg{\,c}\; denotes membership in group $c$ (for CompFS or GroupFS).  
Group indices are assigned only for readability, and the feature order matches Table~\ref{tab:student_features}.}
\label{tab:student_selection}
\end{table*}

\end{document}